\documentclass[runningheads]{llncs}
\usepackage{graphicx}

\usepackage[noend]{algpseudocode}
\usepackage{algorithmicx,algorithm}
\usepackage{booktabs}
\usepackage{amsmath}
\usepackage{multirow}
\usepackage{subfigure}
\usepackage{xcolor}
\usepackage{wrapfig}

\makeatletter
\newcommand{\printfnsymbol}[1]{%
  \textsuperscript{\@fnsymbol{#1}}%
}
\makeatother

\begin{document}
\title{On Robustness and Bias Analysis of BERT-based Relation Extraction}
%

\author{Luoqiu Li\inst{1,2}
\and Xiang Chen \inst{1,2}
\and Hongbin Ye \inst{1,2}
\and Zhen Bi \inst{1,2} 
\and Shumin Deng \inst{1,2}
\and Ningyu Zhang \inst{1,2}\printfnsymbol{1}
\and Huajun Chen \inst{1,2}\thanks{Corresponding author.}
}
\authorrunning{L. Author et al.}
%
\institute{Zhejiang University \& AZFT Joint Lab for Knowledge Engine \and
Hangzhou Innovation Center, Zhejiang University 
\email{\{luoqiu.li,xiang\_chen,yehongbin,bizhen\_zju,231sm,zhangningyu,huajunsir\}@zju.edu.cn}
}

\maketitle              
\begin{abstract}

Fine-tuning pre-trained models have achieved impressive performance on standard natural language processing benchmarks. However, the resultant model generalizability remains poorly understood. We do not know, for example, how excellent performance can lead to the perfection of generalization models. In this study, we analyze a fine-tuned BERT model from different perspectives using relation extraction. We also characterize the differences in generalization techniques according to our proposed improvements. From empirical experimentation, we find that BERT suffers a bottleneck in terms of robustness by way of randomizations, adversarial and counterfactual tests, and biases (i.e., selection and semantic). These findings highlight opportunities for future improvements. Our open-sourced testbed \textbf{DiagnoseRE} is available in \url{https://github.com/zjunlp/DiagnoseRE}.

\end{abstract}
\section{Introduction} 
Self-supervised pre-trained language models (LM), such as the BERT \cite{bert} and RoBERTa \cite{liu2019roberta}, providing powerful contextualized representations, has achieved promising results on standard Natural Language Processing (NLP) benchmarks. However, the generalization behaviors of these types of models remain largely unexplained. 

In NLP, there is a massive gap between task performance and the understanding of model generalizability. Previous approaches indicated that neural models suffered from poor \textbf{robustness} when encountering \emph{randomly permuted contexts} \cite{DBLP:conf/acl/RibeiroWGS20}, \emph{adversarial examples} \cite{jin2019bert}, and \emph{contrastive sets} \cite{DBLP:journals/corr/abs-2004-02709}. Moreover, neural models are susceptible to \textbf{bias} \cite{shah2019predictive}, such as \emph{selection} and \emph{semantic} bias. Concretely, models often capture superficial cues associated with dataset labels which are generally not useful. For example, the term
, ``airport,'' may indicate the output of the relation, ``place\_served\_by\_transport\_hub,'' in the relation extraction (RE) task. However, this is clearly the result of a biased assumption. 
 
Notably, there have been scant studies that analyzed the generalizability of NLP models \cite{fu2020rethinking,DBLP:conf/acl/RibeiroWGS20,li2021normal}. This is surprising because this level of understanding could not only be used to figure out missing connections in state-of-the-art models, but it could also be used to inspire important future studies while forging new ideas. In this study, we use RE as the study case and diagnose its generalizability in terms of robustness and bias. Specifically, we answer five crucial, yet rarely asked, questions about the pre-trained LM BERT \cite{bert}.

\textbf{Q1}: Does BERT really have a generalization capability, or does it make shallow template matches? For this question, we leverage a randomization test for entity and context to analyze BERT's generalizability. Furthermore, we utilize data augmentation to determine whether this is beneficial to generalization. 
\textbf{Q2}: How well does BERT perform with adversarial samples in terms of RE? For this question, we introduce two types of adversarial methods to evaluate its performance. Then, we conduct experiments to understand how adversarial training influences BERT's generalizability. 
\textbf{Q3}: Can BERT generalize to contrast sets, and does counterfactual augmentation help? For this question, we evaluate whether the model can identify negative samples via contrastive sets (samples within a similar context but with different labels). We also propose a novel counterfactual data augmentation method that does not require human intervention to enhance generalization.  
\textbf{Q4}: Can BERT learn simple cues (e.g., lexical overlaps) that work well with most training examples but fail on more challenging ones? We conduct an in-depth analysis and estimate whether its tokens are prone to biased correlations. We also introduce a de-biased method to mitigate selection bias.  
\textbf{Q5}: Does semantic bias in the pre-trained LM hurt RE generalization? We attempt to identify whether these biases exist in BERT, and we introduce an entity-masking method to address this issue. 
 
\textbf{Main Contributions} This paper provides an understanding of BERT's generalization behavior from multiple novel perspectives, contributing to the field from the following perspectives. We first identify the shortcoming of previous RE models in terms of robustness and bias and suggest directions for improvement. Other tasks can benefit from the proposed counterfactual data augmentation method, which notably does not require human intervention. This research also enhances the generalization of two sampling approaches to bias mitigation. We also provide an open-source testbed, ``DiagnoseRE,'' for future research purposes. Ours is the first approach that applies adversarial and counterfactual tests for RE. Our approach can be readily applied to other NLP tasks such as text classification and sentiment analysis. 

\textbf{Observations} We find that BERT is sensitive to random permutations (i.e., entities), indicating that fine-tuning pre-trained models still suffer from poor robustness. We also observe that data augmentation can benefit performance. BERT is found to be vulnerable to adversarial attacks that comprise legitimate inputs that are altered by small and often imperceptible perturbations. Adversarial training can help enhance robustness, but the results are still far from satisfactory. We find that model performance decays in the contrast setting, but counterfactual data augmentation does enhance robustness. BERT is susceptible to learning simple cues, but re-weighting helps to mitigate bias. There exists a semantic bias in the model that hurts generalization, but entity masking can slightly mitigate this. 
 
\section{Related Work}
\subsection{Relation Extraction}
Neural models have been widely used for RE because they accurately capture textual relations without explicit linguistic analyses \cite{zeng2015distant,lin2016neural,zhang2018capsule,ye2020contrastive,yu2020bridging,zhang2020openue}. To further improve their performance, some studies have incorporated external information sources \cite{zeng2016incorporating,ji2017distant,han2018neural} and advanced training strategies \cite{ye2017jointly,liu2017soft,huang2017deep,feng2018reinforcement,zeng2018large,wu2017adversarial,qin2018dsgan,zhang2019long,zhang2020relation}. Leveraging the prosperity of pre-trained LMs, \cite{wang2019extracting} utilized a pre-trained LM for RE: the OpenAI generative pre-trained transformer. \cite{alt2019fine} proposed a solution that could complete multiple entities RE tasks using a pre-trained transformer. Although they achieved promising results on benchmark datasets, the generalizability of RE was not well examined. To the best of our knowledge, we are the first to rigorously study the generalizability of RE. 

\subsection{Analyzing the Generalizability of Neural Networks}
Most existing works \cite{fu2020rethinking} analyzed the generalizability of neural networks using parameters and labels and influencing the training process on a range of classification tasks.
\cite{arpit2017closer} examined the role of memorization in deep learning, drawing connections to capacity, generalization, and adversarial robustness. \cite{fort2019stiffness} developed a perspective on the generalizability of neural networks by proposing and investigating the concept of neural-network stiffness. \cite{zhong2019closer} sought to understand how different dataset factors influenced the generalization behavior of neural extractive summarization models. For NLP, \cite{alt2020probing} introduced 14 probing tasks to understand how encoder architectures and their supporting linguistic knowledge bases affected the features learned by the encoder. \cite{alt2020tacred} attempted to answer whether or not we have reached a performance ceiling or if there was still room for improvement for RE. The current study aims to better understand the generalizability of the fine-tuned pre-trained BERT models regarding robustness and bias. 

\section{Task, Methods, and Datasets}

\begin{table*}[!htbp]
   \fontsize{8}{10}\selectfont
   \centering
   \caption{Outline of our experiment designs.}
\begin{tabular}{cccc}
\toprule
 \textbf{Q.} &  \textbf{Perspectives}& \textbf{Evaluation Settings}& \textbf{Improved Strategies}\\
\midrule
 
 Q1 & Randomization & Random Permutation & Data Augmentation (DA) \\
 Q2 & Adversarial & Adversarial Attack & Adversarial Training (Adv) \\
 Q3 & Counterfactual & Contrastive Masking & Counterfactual Data Augmentation (CDA) \\
\midrule
 Q4 & Selection & Frequent Token Replacement & De-biased Training \\
 Q5 & Semantic & Entity-Only & Selective Entity Masking \\
 \bottomrule
\end{tabular}

  \label{arc}
\end{table*}

\subsection{Task Description}

\quad\textbf{Definition 1.}\textbf{ Robustness} is a measure that indicates whether the model is vulnerable to small and imperceptible permutations originating from legitimate inputs. 

\textbf{Definition 2.} \textbf{Bias} is a measure that illustrates whether the model learns simple cues that work well for the majority of training examples but fail on more challenging ones.  

RE is usually formulated as a sequence classification problem. For example, given the sentence, ``Obama was born in Honolulu,'' with the head entity, ``Obama,'' and the tail entity, ``Honolulu,'' RE assigns the relation label, ``place\_of\_birth,'' to the instance. Formally, let $X=\left\{x_{1}, x_{2}, \ldots, x_{L}\right\}$ be an input sequence, $h,t \in X$ be two entities, and $Y$ be the output relations. The goal of this task is to estimate the conditional probability, $P(Y|X) = P(y|X,h,t)$

\subsection{Fine-tuning the Pre-trained Model for RE}
To evaluate the generalization ability of RE, we leveraged the pre-trained BERT base model (uncased) \cite{bert}. Other strong models (e.g., RoBERTa \cite{liu2019roberta} and XLNet \cite{yang2019xlnet}) could also be leveraged. 

We first preprocessed the sentence, $\mathbf{x}=$ $\{w_1,$ $w_2,$ $h,$ $\dots,$ $t$,...,$w_L\}$, for BERT's input form: $\mathbf{x}=$ $\{$[CLS]$,$ $w_1,$ $w_2,$ [E1], $h,$ [/E1], $\dots,$ [E2], $t,$ [/E2],..., $w_L,$ [SEP]$\}$, where $w_i, i \in  [1, n]$ refers to each word in a sentence and $h$ and $t$ are head and tail entities, respectively. [E1], [/E1], [E2], and [/E2] are four special tokens used to mark the positions of the entities. As we aimed to investigate BERT's generalization ability, we utilized the simplest method, i.e., [CLS] token, as the sentence-feature representation. We used a multilayer perceptron to obtain the relation logits, and we utilized cross-entropy loss for optimization. 

\subsection{RE Datasets for Evaluation}

We conducted experiments on two benchmark datasets: Wiki80 and TACRED. The Wiki80 dataset\footnote{\url{https://github.com/thunlp/OpenNRE}} \cite{DBLP:conf/emnlp/HanZYWYLS18} was first generated using distant supervision. Then, it was filtered by crowdsourcing to remove noisy annotations. The final Wiki80 dataset consisted of 80 relations, each having 700 instances. TACRED\footnote{\url{https://nlp.stanford.edu/projects/tacred/}} \cite{zhang2017position} is a large-scale RE dataset that covers 42 relation types and contains 106,264 sentences. Each sentence in two datasets has only one relation label. To analyze the RE model's generalization, we constructed our robust/de-biased test set based on Wiki80 and TACRED. We evaluate the generalization of BERT on those test sets. More details can be found in the following sections. We used the micro F1 score to evaluate performance.   

\section{Diagnosing Generalization with \emph{Robustness}}

\begin{figure} \centering
  \includegraphics[width=1\textwidth]{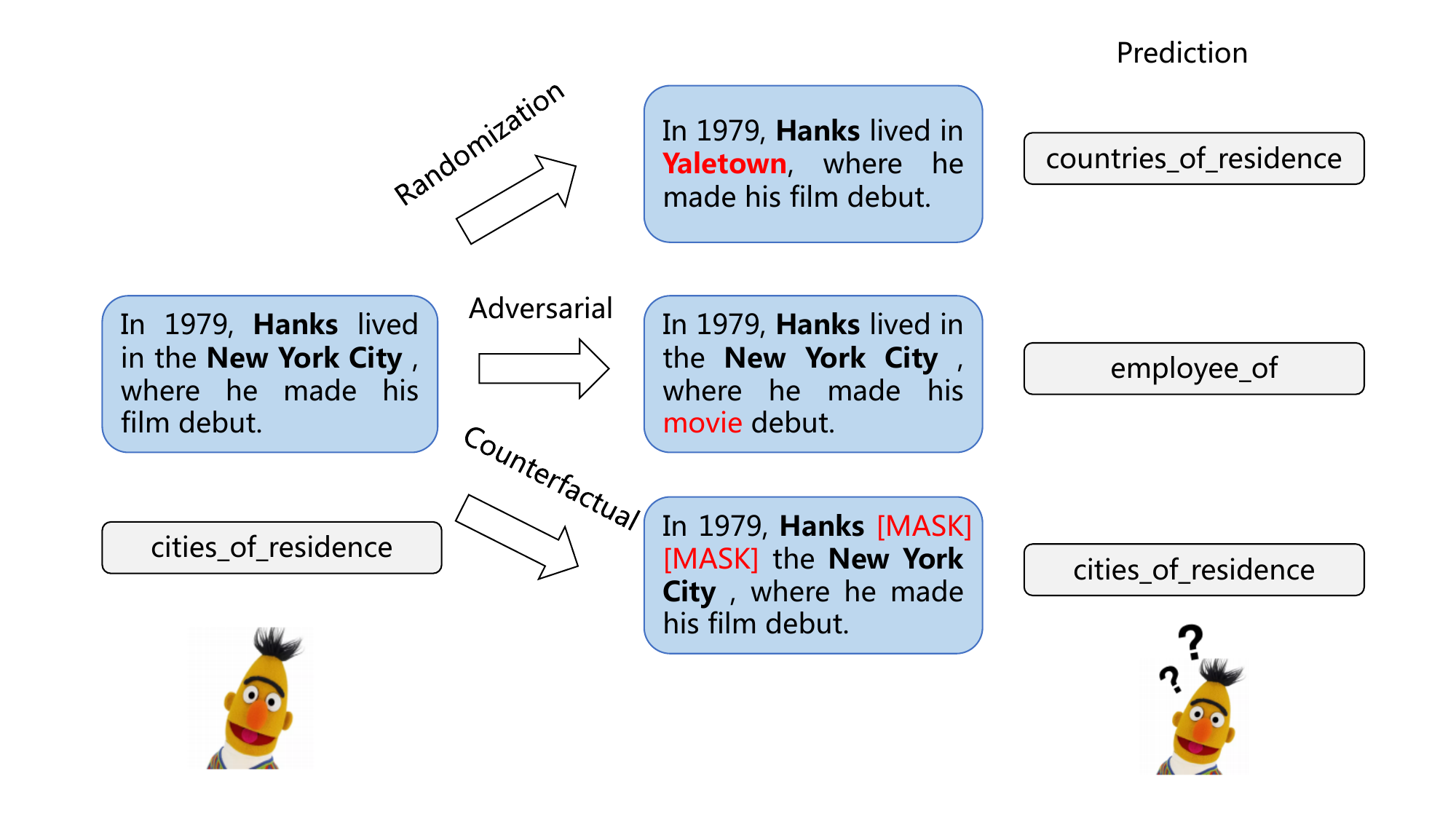}
\caption{Diagnosing generalization with robustness. }
\label{robust}
\end{figure}

We studied the robustness of RE from three aspects: randomized \cite{DBLP:conf/acl/RibeiroWGS20}, adversarial \cite{jin2019bert}, and counterfactual \cite{DBLP:journals/corr/abs-2004-02709}. For example, we would like to investigate the performance of fine-tuned models having diverse surface forms, adversarial permutations, and contrast settings, as shown in Figure \ref{robust}. The randomization test aims to probe the performance with random token permutations. Meanwhile, adversarial analysis is used to study its stability when encountering adversarial permutations. The contrast set is used to analyze whether the model has captured the relevant phenomena, compared with standard metrics from i.i.d. test data. 

\subsection{Randomization Test}\label{diversity}

To conduct the randomization test for RE, we utilized a random permutation for tokens in the test set to construct new robust sets. Note that the entity and context may provide different contributions to the performance of the RE task. Thus, we introduced two types of permutation strategies regarding entity and context as follows:

\textbf{Entity Permutation} is used to investigate the diversity of name entities for RE, which replaces the same entity mention with another entity having the same entity type. Thus, we can identify robust performance with different sentence entities. For example, in Figure \ref{robust}, given the sentence, ``In 1979, \emph{Hanks} lived in the \emph{New York City}, where he made his film debut,'' we replace the entity ``\emph{New York City}'' with the entity ``Yaletown,'' having the same type, ``LOCATION,'' to construct a new testing instance.

\textbf{Context Permutation} is used to investigate the impact of context. Unlike entity permutation, context permutation replaces each word between two entities with similar semantic words. For example, given the sentence, ``Utility permits have been issued to extend a full from Baltimore to \emph{Washington DC}, between Penn Station in Baltimore to \emph{Washington Union Station},'' we replace the word ``Station'' between two entities with a similar semantic word ``Stop.''

\begin{table*}[!htbp]
   \centering
   \caption{Randomization test results from the Wiki80 and TACRED datasets.}
\begin{tabular}{c|c|ccc|c|ccc}
\toprule
& \multicolumn{4}{c|}{\textbf{Wiki80}} & \multicolumn{4}{c}{\textbf{TACRED}} \\
\cline{2-9}
& Origin & \multicolumn{3}{c|}{Robust} &Origin &  \multicolumn{3}{c}{Robust} \\
\cline{2-9}
 Model& All&Entity & Context& All& All &Entity& Context& All\\
 \midrule
 BERT&86.2 &78.4 &81.6 &79.1 &67.5 &57.8 &60.4&58.3 \\
 \midrule
 BERT+DA(Entity)&85.9 &85.5 &85.7 &85.6 &64.3 &64.6 &64.6 &64.6 \\
 BERT+DA(Context)&85.7 &83.2 &85.6 &83.9 &63.6 &61.6 &63.6 &62.1 \\
 BERT+DA(All)&85.7 &85.7 &86.1 &85.8 &63.8 &64.1 &64.4 &64.1 \\
\bottomrule
\end{tabular}

  \label{div_res}
\end{table*}

Specifically, we leveraged the CheckList behavioral testing tool\footnote{\url{https://github.com/marcotcr/checklist}} \cite{DBLP:conf/acl/RibeiroWGS20} to generate entity and context permutations, and we utilized an invariance test (INV) to apply label-preserving perturbations to inputs while expecting the model prediction to remain the same. We leveraged three methods to generate candidate token replacements. First, we used WordNet categories (e.g., synonyms and antonyms). We selected context-appropriate synonyms as permutation candidates. Furthermore, we used additional common fill-ins for general-purpose categories, such as named entities (e.g., common male and female first/last names, cities, and countries) and protected-group adjectives (e.g., nationality, religion, gender, and sexuality) for generating permutation candidates. These two methods can generate vast amounts of robust test instances efficiently. Additionally, we leveraged the pre-trained LM RoBERTa \cite{roberta} to generate permutation candidates. We randomly masked tokens in the sentences and generated the mask token via a mask LM. For example, ``New York is a [MASK] city in the United States'' yields \{``small", ``major", ``port", ``large"\}. We randomly selected the top-two tokens as permutation candidates and leveraged three strategies for robust set construction. 

To ensure that the random token replacement was label-preserving, we manually evaluated the quality of the generated instances. We randomly selected 200 instances and found that only two permuted sentences had the wrong labels, indicating that our robust set was of high quality. In total, we generated  5,600/15,509 test instances of entity and context permutations on both datasets. 
We constructed a combined robust set (Table \ref{div_res}) with both entity and context permutations. We evaluated the performance of BERT with the original test set and the robust set. We also trained BERT with data augmentation regarding entity (BERT+DA(Entity)) and context (BERT+DA(Context)) for evaluation. We employed Adam~\cite{Kingma2015AdamAM}, and the initial learning rate was 2e-5. 
The batch size was 32, and the maximum epoch was 5.
The hyperparameter was the \textbf{same} for different experiments. 

\textbf{Results and Analysis.}
From Table \ref{div_res}, we observe that the overall performance decayed severely in the robust set of entity and context permutations. BERT had a more remarkable performance decay with entity permutations, which indicates that the model was unstable with different head and tail entities. Furthermore, we found that BERT+DA achieved better performance in both the original and robust test sets. However, the robust test set's overall performance was still far from satisfactory. Thus, more robust algorithms are needed for future studies. 

\subsection{Adversarial Testing}\label{adv}
In this section, we focus on the problem of generating valid adversarial examples for RE and defending from adversarial attacks with adversarial training. Given a set of $N$ instances, $\mathcal{X}= \{X_1, X_2,\dots, X_N\}$ with a corresponding set of labels, $\mathcal{Y}=\{Y_1, Y_2,\dots, Y_N$\}, we have a RE model, $ \mathcal{Y}  = RE(\mathcal{X})$, which  is trained via the input $\mathcal{X}$ and $\mathcal{Y}$.

The adversarial example $X_{\mathrm{adv}}$ for each sentence $X \in \mathcal{X}$  should conform to the requirements as follows: 
\begin{equation}
RE\left(X_{\mathrm{adv}} \right) \neq RE(X), \text { and } \operatorname{Sim}\left(X_{\mathrm{adv}}, X\right) \geq \epsilon,
\label{eq:ad_requirement}
\end{equation}
where $\mathrm{Sim}$ is a similarity function  and $\epsilon$ is the minimum similarity between the original and adversarial examples. In this study, we leveraged two efficient adversarial attack approaches for RE: PWWS \cite{ren2019generating} and HotFlip \cite{ebrahimi2017hotflip}.  

\textbf{PWWS}, a.k.a., Probability Weighted Word
Saliency, is a method based on synonym replacement. PWWS firstly find the corresponding substitute based on synonyms or entities and then decide the replacement order.  Specifically, given a sentence of $L$ words, $X=\left\{w_{1}, w_{2}, \ldots, w_{L}\right\}$, we first selected the important prediction tokens having a high score of $I_{w_{i}}$, which is calculated as the prediction change before and after deleting the word.  Then, we gathered a candidate set with the WordNet synonyms and named entities. To determine the priority of words for replacement,  we score each proposed substitute word $w_{i}^{*}$ by evaluating the $i^{th}$  value of $\mathbf{S}(\mathbf{x})$.
The score function $H\left(\mathbf{x}, \mathbf{x}_{i}^{*}, w_{i}\right)$ is defined as:
\begin{equation}
    H\left(\mathbf{x}, \mathbf{x}_{i}^{*}, w_{i}\right)=\phi(\mathbf{S}(\mathbf{x}))_{i} \cdot P\left(y_{\text {true }} \mid \mathbf{x}\right)-P\left(y_{\text {true }} \mid \mathbf{x}_{i}^{*}\right)
    \label{pwws}
\end{equation}
where $\phi(\mathbf{z})_{i}$ is the softmax function,  Eq. \ref{pwws}  determines the replacement order.  Based on $H\left(\mathbf{x}, \mathbf{x}_{i}^{*}, w_{i}\right)$,  all the words $w_i$ in $X$ are sorted in descending order. We then use each word $w_i$ under this order. Specifically, we greedily select the substitute word $w_{i}^{*}$ for $w_i$ to be replaced  which can make the final classification label change iteratively through the process until enough words have been replaced.


\textbf{HotFlip} is a gradient-based method that generates adversarial examples using character substitutions (i.e., ``flips''). HotFlip also supports insertion and deletion operations by representing them as sequences of character substitutions. It uses the gradient from the one-hot input representation to estimate which individual change has the highest estimated loss efficiently. Further, HotFlip uses a beam search to find a set of manipulations that work together to confuse a classifier. 

We conducted experiments based on OpenAttack\footnote{\url{https://github.com/thunlp/OpenAttack}} and generated adversarial samples of PWWS and HotFlip to construct a robust test set separately. We also conducted adversarial training experiments \cite{tramer2017ensemble} to improve the robustness of machine-learning models by enriching the training data using generated adversarial examples. We evaluated the performance of the vanilla BERT and the version using adversarial training (BERT+Adv) in both the original and robust test sets. 

\begin{table}[!htbp]
   \centering
   \caption{Adversarial test results from the Wiki80 and TACRED datasets. The former indicates the results of adversarial sets while the latter indicates the results of orginal sets.}
\begin{tabular}{c|c|c}
\toprule
Model& \textbf{Wiki80} & \textbf{TACRED}\\
 \midrule
 BERT (Origin)&86.2 &67.5 \\
 \midrule
 BERT (PWWS/Origin)&52.9/86.2&37.7/67.5\\
 BERT (HotFlip/Origin)&56.3/86.2&49.2/67.5 \\
 \midrule
 BERT+Adv (PWWS/Origin)&86.4/86.4&72.1/65.9\\
BERT+Adv (HotFlip/Origin)&87.0/86.6&73.9/67.8\\
 \bottomrule
\end{tabular}

  \label{adv_res}
\end{table}

\textbf{Results and Analysis.} From Table \ref{adv_res}, we observe that BERT achieved significant performance decay with PWWS and HotFip, revealing that fine-tuned RE models are vulnerable to adversarial attacks. We noticed that adversarial training helped achieve better performance. Conversely, the original set's evaluation results were slightly decayed, as was also found in \cite{wen2019towards}. We, therefore, argue that there is a balance between adversarial and original instances, and more reasonable approaches should be considered.

\begin{table*}
   \fontsize{6}{10}\selectfont 
   \centering
   \caption{Important token generation with attention, integrated gradients and contrastive masking.}
\begin{tabular}{c|p{7cm}|c}
\toprule
Method& Text & Label\\
\midrule
Attention& \colorbox[rgb]{0.753,0.282,0.318}{\color{black}[CLS]} In 1979, Hanks \colorbox[rgb]{0.96,0.76,0.76}{\color{black}lived} in the New \colorbox[rgb]{0.753,0.282,0.318}{\color{black}York City} ,  ... [SEP]& cities\_of\_residence \\
 Integrated Gradients& [CLS] In 1979, Hanks \colorbox[rgb]{0.753,0.282,0.318}{\color{black}lived} \colorbox[rgb]{0.753,0.282,0.318}{\color{black}in} the \colorbox[rgb]{0.96,0.76,0.76}{\color{black}New York City}, ... [SEP]& cities\_of\_residence \\
 \midrule
  Constasive Masking& [CLS] In 1979, Hanks [MASK]  [MASK] the New York City , ... [SEP]& \textbf{NOT}\_cities\_of\_residence \\
 \bottomrule
\end{tabular}

  \label{att}
\end{table*}

\subsection{Counterfactual Test}\label{cf}
Previous approaches \cite{DBLP:journals/corr/abs-2004-02709,zhou2020can} indicated that fine-tuned models, such as BERT, learn simple decision rules that perform well on the test set but do not capture a dataset’s intended capabilities. For example, given the sentence ``In 1979, Hanks lived in New York City, where he ...,'' we can classify the sentence into the label, ``cities\_of\_residence,'' owing to the phrase ``lived in.'' We seek to understand whether the prediction will change, given a sentence lacking such an indicating phrase. This sentence indicates the contrast set, as noted in \cite{gardner2020evaluating}. We humans can easily identify that this instance does not contain the relation ``cities\_of\_residence,'' via counterfactual reasoning. Motivated by this, we took our first step toward analyzing the generalization of RE in contrast sets. In this setting, we should generate examples of few permutations but with opposite labels.
In contrast to the previous approach \cite{gardner2020evaluating}, which utilized crowdsourcing, we generated the contrast set automatically. Hence, we proposed a novel counterfactual data augmentation method lacking human intervention to generate contrast sets. We first generated the most informative tokens of the sentences regarding its relation labels, and we then introduced contrastive masking to obscure those tokens to generate the contrast set, as shown in Table \ref{att}.

Specifically, we leveraged {\it integrated gradients} \cite{sundararajan2017axiomatic} to generate informative tokens. We did not leverage attention scores \cite{wiegreffe2019attention} because \cite{weakAttention} pointed out that analyzing only attention weights would be insufficient when investigating the behavior of the attention head. Furthermore, attention weights disregarded the hidden vector's values. Moreover, as shown in Table \ref{att}, we empirically observed that attention scores were not suitable for generating important tokens. 

Intuitively, integrated gradients is a variation on computing the gradient of the prediction output w.r.t. features of the input, which simulate the process of pruning the specific attention head from the original attention weight, $\alpha$, to a zero vector, $\alpha' $, via back-propagation \cite{zhou2020can}.  
Note that integrated gradients can generate attribution scores reflecting how much changing the attention weights will change the model's outputs. In other words, the higher of attribution score, the greater importance given to attention weights. 
Given an input, $x$, the attribution score of the attention head, $t$, can be computed using:
\begin{equation}
\small
Atr(\alpha^t) = (\alpha^t - \alpha'^t) \otimes \int_{x=0}^{1} \frac{\partial F(\alpha' + x (\alpha - \alpha'))}{\partial \alpha^t} dx, 
\end{equation}
where $\alpha=[\alpha^1,\dots,\alpha^{T}]$, and $\otimes$ is the element-wise multiplication. $Atr(\alpha^t) \in \mathbf{R}^{n \times n}$ denotes the attribution score, which corresponds to the attention weight $\alpha^t$. Naturally, $F(\alpha^{\prime} + x(\alpha - \alpha^{\prime}))$ is closer to $F(\alpha^{\prime})$ when $x$ is closer to $0$, and it is closer to $\alpha$ when $x$ is closer to $1$. 
We set the uninformative baseline $\alpha^{\prime}$ as a zero vector and denote  $Atr(\alpha^t_{i,j})$ as the interaction from token $\mathbf{h}_i$ to $\mathbf{h}_j$. 
We approximate $Atr(\alpha^t)$ via a gradient summation function following \cite{ig,zhou2020can}:
\begin{equation}
\small
    Atr(\alpha^t):: = (\alpha^t - \alpha'^t) \odot  \sum_{i=1}^s \frac{\partial F(\alpha' + i/s (\alpha - \alpha'))}{\partial \alpha'^t} \times \frac{1}{s}, 
\end{equation}
where $s$ is the number of approximation steps for computing the integrated gradients. We selected the top $k=1,2$ informative tokens from instance and implemented contrastive masking by replacing informative tokens with unused tokens (e.g., [unused5]). We leveraged this procedure for both training and testing datasets, and the overall algorithm is:
 
\begin{algorithm}[th]
\begin{algorithmic}[1]
\caption{Counterfactual Data Augmentation for RE} 
 \State Train Relation Classifier $RE$ with X,Y
\For{$x$ in $X$}

\State ig = IntegratedGradients($RE$)

\State attributes = ig.attribute(X)

\State candidates = select\_top\_k(attributes)

\State $x^{mask}$ = mask(x,candidates)

\State X $\leftarrow$ X $\cap$ $x^{mask}$ Y $\leftarrow$ Y $\cap$ NA

\EndFor

 \State Re-train $RE$ with X,Y
\label{alg} 
\end{algorithmic}
\end{algorithm}

We generated 15,509 samples to construct a robust contrast set. Since the Wiki80 dataset does not contain NA relation, we only evaluate results on the TACRED dataset with the performance of vanilla BERT and BERT with counterfactual data augmentation (BERT+CDA). Note that the contrast set of RE comprised instances with \textbf{NOT\_such\_relation (NA)} labels. Thus, we utilized the F1 score including NA. 

\begin{table}[!htbp]
   \centering
   \caption{Counterfactual analysis results on the TACRED dataset (F1 score including \textbf{NA}). The former and the latter indicates the results of contrast and  original sets, separately.}
\begin{tabular}{c|c}
\toprule
Model&  \textbf{TACRED} \\
 \midrule
 BERT (Origin) &87.7 \\
 \midrule
BERT (Contrast Set,$k=1$) &32.6 \\ 
BERT (Contrast Set,$k=2$)  &45.1\\
 \midrule
BERT+CDA (Contrast Set/Origin,$k=1$)&89.0/86.8
\\
BERT+CDA (Contrast Set/Origin,$k=2$)&95.0/87.2
\\
 \bottomrule
\end{tabular}

  \label{cf_res}
\end{table}

\textbf{Results and Analysis.} From Table \ref{cf_res}, we notice that BERT achieved poor performance on the robust set, which shows that fine-tuned models lack the ability of counterfactual reasoning. We also found that BERT+CDA achieved better results than BERT on a robust set, indicating that counterfactual data augmentation was beneficial. Note that, unlike previous counterfactual data augmentation approaches, such as \cite{wiegreffe2019attention,chen2020counterfactual}, our method was a simple, yet effective, automatic algorithm that can be applied to other tasks (e.g., event extraction \cite{deng2020meta}, text classification \cite{deng2020low,zhang2020conceptualized}, sentiment analysis \cite{peng2020knowing} and question answering \cite{zhong2020iteratively}).

\section{Diagnosing Generalization with \emph{Bias}}

\begin{figure*} \centering
\includegraphics[width=0.9\textwidth]{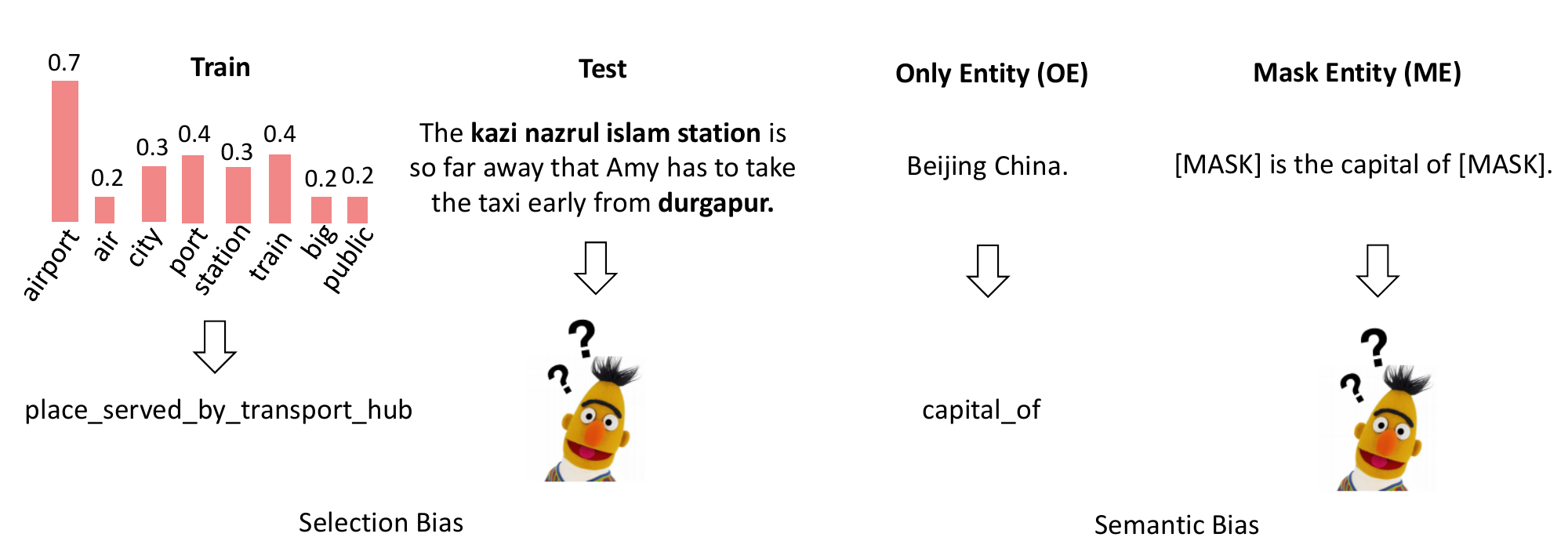}
\caption{Diagnosing generalization with bias.}
\label{bias}
\end{figure*}
\subsection{Selection Bias}\label{datasetbias}
Selection bias emerges from the non-representative observations, such as when the users generating the training observations have different distributions than that in which the model is intended to be applied \cite{shah2019predictive}. For a long time, selection bias (a.k.a. sample bias) has been a concern in social sciences, so much so that considerations of this bias are now considered primary considerations in research design. For the RE, we are the first to have studied selection bias. Given a running example, as shown in Figure \ref{bias}, owing to the high frequency of the token ``airport,'' the fine-tuned model can memorize the correlation between the existence of the token and the relation ``place\_served\_by\_transport\_hub'' by neglecting the low-frequency words (e.g., ``train station'').

The origin of the selection bias is the non-representative data. The predicted output is different from the ideal distribution, for example, because the given demographics cannot reflect the ideal distribution, resulting in lower accuracy. To analyze the effect of selection bias for RE, we constructed a de-biased test set that replaced high-frequency tokens with low-frequency ones. We evaluated the performance of BERT on a de-biased set, and we introduced a simple method, BERT+De-biased, which masks the tokens based on the token frequency (neglecting the stop words and common words, such as ``the,'' ``when,'' and ``none''). 

\begin{table}[!htbp]
   \centering
   \caption{Selection bias analysis results on the Wiki80 and TACRED datasets. The former indicates the results of de-biased sets while the latter indicates the results of orginal sets.}
\begin{tabular}{c|c|c}
\toprule
Model&  \textbf{Wiki80} &\textbf{TACRED}  \\
 \midrule
BERT (Origin)& 86.2& 67.5\\
\midrule
BERT (De-biased) & 80.2
&64.7
\\
BERT+De-biased (De-biased) &84.3/85.3
&67.1/67.4
\\
\bottomrule
\end{tabular}

 \label{selection}
\end{table}
 
\textbf{Results and Analysis.} From Table \ref{selection}, we notice that BERT achieved poor performance on the de-biased set, which shows that there exists a selection bias for RE in previous benchmarks. We also find that BERT+Re-weighting achieved relatively better results, compared with BERT on the de-biased set, indicating that frequent-based re-sampling was beneficial. Note that previous benchmarks (e.g., Wiki80 and TACRED) did not reveal the real data distribution for RE. Therefore, we argue that selection bias may be worse in a real-world setting, and more studies are required.

 \begin{table*}[!htbp]
\centering
\caption{Semantic bias analysis results on the Wiki80 and TACRED dataset.}
\begin{tabular}{c|cccc|cccc}
\toprule
& \multicolumn{4}{c|}{\textbf{Wiki80}} & \multicolumn{4}{c}{\textbf{TACRED}} \\
\cline{2-9}
 Model& Origin&OE&ME&De-biased& Origin&OE&ME&De-biased  \\
 \midrule
 BERT&86.2 &66.5 &52.4 &67.0 &67.5&42.9 &36.6 &57.4 \\
 \midrule
 BERT+ME (50\%)&86.4 &65.3 &73.5 &67.9 &67.9 &42.9 &55.8 &61.5 \\
 BERT+ME (100\%)&86.0 &62.9 &74.7 &68.1 &67.4 &43.6 &55.9 &60.5 \\
 BERT+ME (Frequency)&-&-&-&-&67.9 &40.8 &53.5 &61.6 \\
 \bottomrule
\end{tabular}

  \label{semantic}
\end{table*}

\subsection{Semantic Bias}\label{lmbias}
Embeddings (i.e., vectors representing the meanings of words or phrases) have become a mainstay of modern NLP, which provides flexible features that are easily applied to deep machine learning architectures. However, previous approaches \cite{nadeem2020stereoset} indicate that these embeddings may contain undesirable societal and unintended stereotypes (e.g., connecting medical nurses more frequently to female pronouns than male pronouns). This is an example of semantic bias.

The origin of the Semantic bias may be the parameters of the embedding model. Semantic bias will indirectly affect the outcomes and error disparities by causing other biases (e.g., diverging word associations within embeddings or LMs\cite{shah2019predictive}). To analyze the effects of semantic bias, we conducted experiments with two settings, as inspired by \cite{han2020more}: a \emph{masked-entity} (ME) setting, wherein entity names are replaced with a special token, and an \emph{only-entity} (OE) setting, wherein only the names of the two entities are provided. We also constructed a de-biased test set in which instances were wrongly predicted in the OE setting. We conducted experiments on these datasets and introduced a simple method of selective entity masking to mitigate semantic bias. We masked $K$\% of the entities with unused tokens to guide the model to pay closer attention to the context (BERT+ME ($K$\%)). We intuitively selected $K$ via entity-pair frequencies (BERT+ME (Frequency))\footnote{On Wiki80, the entity pair is unique. Thus, we do not conduct BERT+ME (Frequency)}. 
 
\textbf{Results and Analysis.} From Table \ref{semantic}, we observe that the models suffered a significant performance drop with both the ME and OE settings. Moreover, it was surprising to notice that, in most cases, with only entity names can archive better performance than those of text only with entities masked. These empirical results illustrate that both entity names and text provided important information for RE, and entity names contributed even more, indicating the existence of semantic bias. It is contrary to human intuition since we identity relations mainly through the context between the given entities, whereas the models take more entity names into consideration. Furthermore, we noticed that BERT achieved poor performance on the de-biased set. We also found that BERT+ME(k) obtained better performance than BERT and BERT+ME(frequency) achieved the best results on the de-biased set, indicating that selective entity masking was beneficial.

\section{Discussion and Limitation}
\begin{wrapfigure}{r}{0.5\textwidth}
\centering
\includegraphics[width=0.45\textwidth]{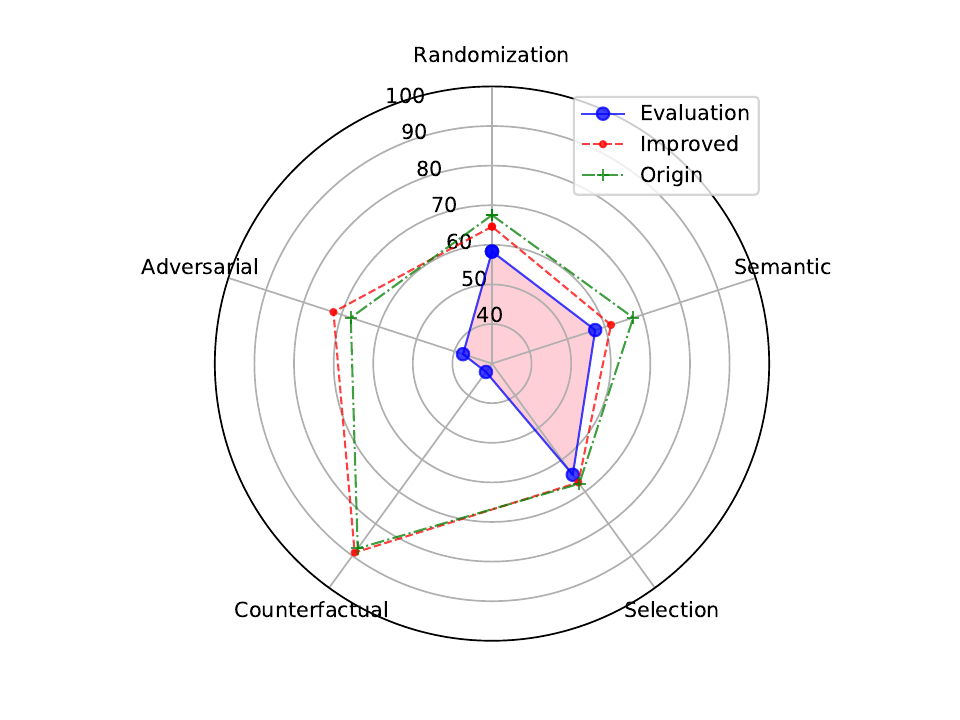}
\caption{Generalization analysis results of RE on TACRED. The \emph{origin} and \emph{evaluation} refer to the BERT performance on the original, robust/de-biased test set, respectively. The \emph{improved} indicates the performance of our proposed methods.}
\label{gen}
\end{wrapfigure}

\textbf{Evaluation of NLP models.} Several models that leveraged pre-trained and fine-tuned regimes have achieved promising results with standard NLP benchmarks. However, the ultimate objective of NLP is generalization. Previous works \cite{DBLP:conf/acl/RibeiroWGS20} attempted to analyze this generalization capability using NLP models' comprehensive behavioral tests. Motivated by this, we took the CheckList paradigm a step further to investigate generalization via robustness and bias. We used RE as an example and conducted experiments. Empirically, the results showed that the BERT performed well on the original test set, but it exhibited poor performance on the robust and de-biased sets, as shown in Figure \ref{gen}. This indicates that generalization should be carefully considered in the future. 

\textbf{Limitations.} We only considered single-label classifications because there was only one relation for each instance. Arguably, there could exist multiple labels for each instance (e.g., multiple RE \cite{zeng2020copymtl}). Moreover, apart from selection and semantic biases, label bias exists with the overamplification of NLP \cite{shah2019predictive}. Using the label bias as an example, the distribution of the dependent variable in the train set may diverge substantially from the test, leading to the deterioration of performance. We leave this problem for future work.

\section{Conclusion and Future Work}
We investigated the generalizability of fine-tuned pre-trained models (i.e., BERT) for RE. Specifically, we diagnosed the bottleneck with regard to existing approaches in terms of robustness and bias, resulting in several directions for future improvement. We introduced several improvements, such as counterfactual data augmentation, sample re-weighting, which can be used to improve generalization. We regard this study as a step toward a unified understanding of generalization, and this offers hopes for further evaluations and improvements of generalization, including conceptual and mathematical definitions of NLP generalization.
\bibliography{aaai21}

 \bibliographystyle{splncs04}
%

\end{document}